\documentclass{article}

\PassOptionsToPackage{numbers,compress}{natbib}
\usepackage[dblblindworkshop,final]{neurips_2026}

\workshoptitle{On-Device Intelligence: Foundation Models under Real-World Constraints}
\makeatletter
\renewcommand{\@trackname}{
NeurIPS 2026 Workshop on On-Device Intelligence: Foundation Models under Real-World Constraints.
}
\makeatother
\usepackage[utf8]{inputenc}
\usepackage[T1]{fontenc}
\usepackage{hyperref}
\hypersetup{hidelinks}
\usepackage{url}
\usepackage{booktabs}
\usepackage{amsmath}
\usepackage{amsfonts}
\usepackage{nicefrac}
\usepackage{microtype}
\usepackage{graphicx}
\usepackage{xcolor}
\usepackage{makecell}
\usepackage{multirow}

\title{Robust Parameter-Efficient LLM Adaptation on Analog Hardware}
\author{
  Jindan Li \\
  Cornell University \\
  \texttt{jl4767@cornell.edu}
  \And
  Zhaoxian Wu \\
  Cornell University \\
  \texttt{zw868@cornell.edu}
  \And
  Tianyi Chen \\
  Cornell University \\
  \texttt{tianyi.chen@cornell.edu}
}

\newcommand{\qametric}[4]{\makecell[c]{#1{\scriptsize$\pm$#2} / #3{\scriptsize$\pm$#4}}}

\definecolor{pennblue}{cmyk}{1,0.65,0,0.30}
\definecolor{pennred}{cmyk}{0,1,0.65,0.34}
\definecolor{mygreen}{rgb}{0.10,0.50,0.10}

\begin{document}
\maketitle

\begin{abstract}
Analog in-memory computing is a promising platform for on-device execution of
large language models because it performs matrix--vector multiplications
(MVMs) in memory and in parallel, reducing data movement. However, limited
digital-to-analog converter precision, input noise, and finite conductance
states can degrade model accuracy, while full-model retraining to address
these effects can be costly. We develop an optimizer-agnostic, parameter-efficient adaptation method
based on Low-Rank Adaptation (LoRA), keeping the pretrained weights stored
on analog arrays fixed while training the LoRA weights to adapt
to downstream tasks and hardware non-idealities. Reliable adaptation
requires handling errors in both forward and backward MVMs and physical
weight updates. We use input reshaping to reduce input-induced MVM errors
and update accumulation to retain small updates before programming them to
finite-state analog devices. Across Llama-3.2-1B-Instruct and Llama-3-8B with both Muon and AdamW, input reshaping improves analog LoRA fine-tuning under noisy MVM computation. Update accumulation separately preserves sub-threshold updates and substantially improves adaptation under finite-resolution programming, including configurations with as few as 20  conductance states. Additional experiments  show consistent held-out negative log-likelihood improvements across noisy analog settings.

\end{abstract}

\section{Introduction}

Large language models (LLMs) rely heavily on matrix--vector multiplications
(MVMs) in their linear layers
\cite{vaswani2017attention,pope2023efficiently}. Analog in-memory computing
(AIMC) \cite{ielmini2018memory,sebastian2020memory} is an attractive hardware
platform for these operations because weights can be stored directly in memory
and MVMs can be performed in place and in parallel
\cite{gokmen2016acceleration,ambrogio2018equivalent,nandakumar2020mixed}.
This reduces data movement and can improve energy efficiency and latency
\cite{horowitz20141,nandakumar2020mixed,rasch2023hardware}, making AIMC a
promising candidate for on-device intelligence.

However, deploying pretrained LLMs on analog hardware can degrade model
accuracy because limited digital-to-analog converter (DAC) precision, input
noise, and finite conductance states introduce errors into the model's
computation. Hardware-aware retraining can mitigate this degradation by
accounting for analog non-idealities during training
\cite{rasch2023hardware,le2023using,buchel2024aihwkit}. However, retraining
the full model requires gradients and optimizer states for all trainable
weights, as well as computation to update those weights. Repeating this
process for downstream tasks can therefore
incur substantial training-memory and computation costs. These costs motivate parameter-efficient fine-tuning (PEFT) for on-device
analog adaptation \cite{ding2023parameter}. We consider Low-Rank Adaptation (LoRA) \cite{hu2022lora,li2026efficient}
to adapt the model to downstream tasks and compensate for hardware-induced
degradation while keeping the pretrained weights fixed. In our setting, the deployed pretrained weights and the
corresponding LoRA weights are executed on analog arrays, with digital support
for input reshaping and update accumulation.

\begin{figure}[t]
  \centering
  \includegraphics[width=\linewidth]{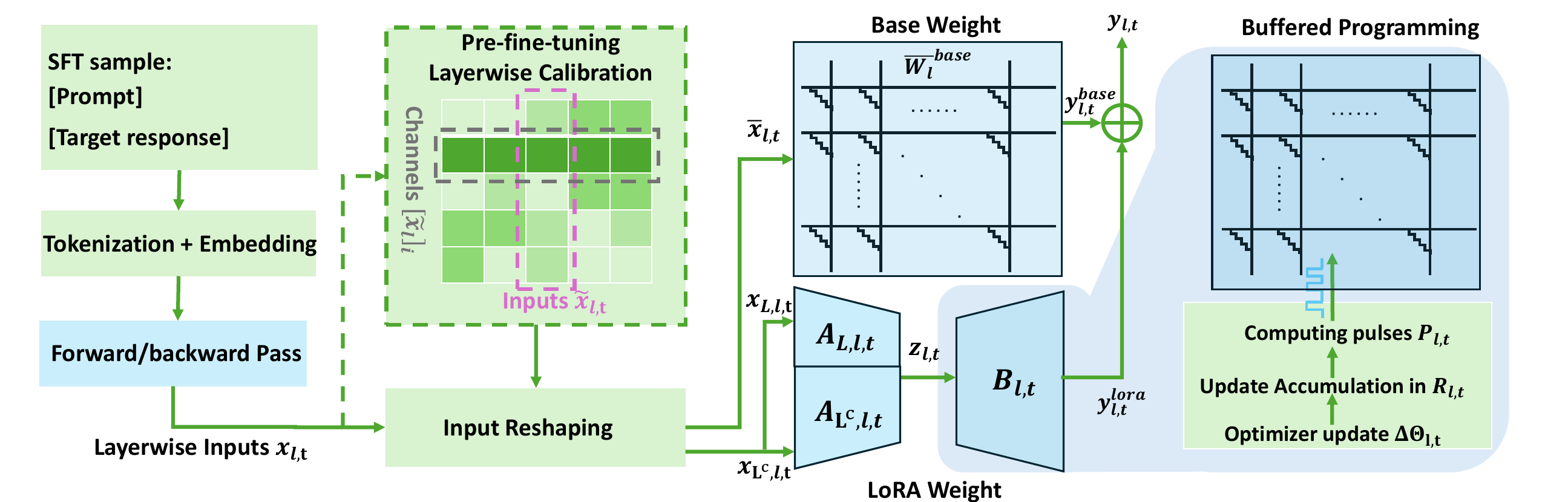}
  \caption{\textbf{Overview of analog LoRA fine-tuning.}
  Pre-fine-tuning calibration provides channel statistics for input reshaping.
  For the frozen base path, function-preserving base-path rescaling reduces the
  input range. For the trainable LoRA path, outlier-channel splitting routes persistent outlier channels through the protected analog path and keeps the remaining channels on the noisy analog path. During weight programming, digital mini-batch
  aggregation combines sample gradients, and the update accumulator stores
  sub-threshold updates until they can be converted into programming pulses.
  Blue and green denote analog and digital computation, respectively.}
  \label{fig:pipeline}
\end{figure}

Reducing the number of trainable weights does not remove the errors
introduced by analog computation and programming. Limited DAC precision and
input noise perturb the forward and backward MVMs used to compute LoRA
gradients. Finite conductance states and non-ideal programming also limit how
accurately the resulting updates can be written to the LoRA weights
\cite{gokmen2016acceleration,gokmen2020algorithm,le2023using}.
Our input reshaping treats the fixed pretrained weights and trainable LoRA
weights differently. For the pretrained weights, function-preserving base-path
rescaling jointly rescales the weights and their inputs to reduce the input
range without changing the ideal computation \cite{xiao2023smoothquant}.
For the LoRA weights, we route persistent input-outlier channels through a protected perfect-I/O analog path, so that their large values no
longer set the quantization scale for the remaining analog inputs. For weight
programming, we use mixed-precision update stabilization: digital mini-batch
aggregation combines gradients before the optimizer step, and update
accumulation retains updates smaller than a conductance step until they can be
programmed \cite{nandakumar2020mixed}.
Figure~\ref{fig:pipeline} summarizes the method.

Our contributions are threefold. First, we develop a lightweight adaptation
method for LLMs on analog hardware, motivated by the needs of on-device
intelligence. The method keeps pretrained weights fixed and trains only a small
set of LoRA weights to address hardware-induced
degradation. Second, we address noisy MVM computation through input reshaping
and limited weight-update resolution through mixed-precision update
stabilization using digital mini-batch aggregation and update accumulation. 
Third, we evaluate the optimizer-agnostic design in simulation on
Llama-3.2-1B-Instruct and Llama-3-8B using Muon
\cite{jordan2024muon,liu2025muon} and AdamW
\cite{loshchilov2017decoupled}. Input reshaping remains effective with
both optimizers under noisy MVM computation, while its gains largely
disappear when MVM input perturbations are removed. Update accumulation separately improves adaptation under finite-resolution weight programming. Additional model and task results show consistent
held-out negative log-likelihood improvements under noisy analog settings.

\section{Related Work}
\noindent\textbf{Parameter-efficient fine-tuning.}
Parameter-efficient fine-tuning methods include Adapter Tuning
\cite{houlsby2019parameter}, Prefix Tuning \cite{li2021prefix}, Prompt Tuning
\cite{lester2021power}, LoRA \cite{hu2022lora}, QLoRA
\cite{dettmers2023qlora}, AdaLoRA \cite{zhang2023adalora}, and DoRA
\cite{liu2024dora}. Recent work further revisits LoRA architectures, optimization, and deployment
considerations from a signal-processing perspective~\cite{li2026low}.
For analog hardware adaptation, prior work keeps the pretrained weights fixed
on AIMC tiles while performing the LoRA computation on separate digital
processors~\cite{li2026efficient}. Its deployment therefore requires a
dedicated digital LoRA execution path whose latency must be balanced with the
AIMC tiles. We instead place the main LoRA computation and
weight updates on analog arrays. This removes the need to rely on a fully
digital LoRA computation path, but makes the trainable LoRA weights themselves
subject to noisy MVMs and non-ideal programming, which motivates the robustness
techniques studied in this work.

\noindent\textbf{Analog training and low-precision computation.}
Analog in-memory training is affected by noisy MVMs, finite conductance states \cite{li2025memory},
peripheral quantization, and non-ideal programming responses \cite{wu2024towards, wu2025analog, xiao2026dynamicsymmetricpointtracking}. Hardware-aware
training and simulation frameworks model these effects during training
\cite{rasch2023hardware,le2023using,buchel2024aihwkit}, while mixed-precision
computational-memory training uses high-precision digital accumulation to
improve weight updates on low-precision analog devices
\cite{nandakumar2020mixed}. 
Recent work further co-designs weight mapping and 
optimization to improve the scalability of
analog training under hardware non-idealities \cite{wu2026makinganalogtrainingscale}.
Complementary work on low-precision transformers
addresses large activation ranges and persistent outliers through
activation--weight rescaling \cite{xiao2023smoothquant}, outlier suppression
\cite{wei2022outlier}, and activation-aware quantization
\cite{lin2024awq}. Noise-optimized rescaling (NoRA) further applies activation--weight
rescaling to analog inference to reduce input-side non-idealities
\cite{hou2025nora}.

\section{Analog LoRA Fine-Tuning under Hardware Non-Idealities}
\label{sec:method}
This section presents our analog LoRA fine-tuning method under non-ideal MVM
computation and physical weight programming. We first define the analog LoRA
computation and the hardware-noise model used during training. We then introduce
input reshaping to reduce DAC-side MVM perturbations, followed by digital
mini-batch aggregation and update accumulation to stabilize finite-state
programming of the trainable LoRA weights.
\subsection{Analog LoRA and non-ideal MVMs}

We start from a pretrained transformer and map part of the pretrained weights
and the corresponding trainable LoRA weights to analog arrays. For a layer with input dimension $n$ and output dimension $m$, the pretrained
base weights
$\mathbf W^{\mathrm{base}}\in\mathbb R^{m\times n}$ are programmed once and
kept fixed, while the LoRA matrices
$\mathbf A\in\mathbb R^{r\times n}$ and
$\mathbf B\in\mathbb R^{m\times r}$ are updated during fine-tuning.
Both the pretrained and LoRA weights are used in non-ideal analog forward and
backward MVMs. Because the LoRA weights are trainable, they are additionally
subject to finite-state and non-ideal physical programming when 
updates are written to the arrays.
For input $\mathbf x\in\mathbb R^n$, the ideal output $\mathbf y\in\mathbb R^m$ is
\begin{equation}
\label{eq:lora_forward}
   \mathbf y
    =
    \mathbf W^{\mathrm{base}} \mathbf x
    +
    \frac{\alpha}{r}\mathbf B\mathbf A\mathbf x ,
\end{equation}
where $r$ is the LoRA rank and $\alpha$ is the LoRA scaling factor.

Our input reshaping is motivated by the persistent activation outliers observed
in pretrained transformers \cite{xiao2023smoothquant,lin2024awq}. We use absolute-maximum input scaling, which normalizes each MVM
input by its maximum absolute value $\|\mathbf x\|_\infty$ so that the largest
input magnitude determines the DAC range. As a result, a few persistent
outlier channels can repeatedly enlarge the quantization step for all remaining
channels. We therefore reshape the MVM inputs to reduce this dynamic range while
preserving the intended computation.
For a general weight $\mathbf W$, we model a non-ideal analog MVM as
\(
f_{\mathrm{out}}\!\left(
    \mathbf W f_{\mathrm{DAC}}(\mathbf x)
\right),
\)
where $f_{\mathrm{DAC}}(\cdot)$ captures input-side conversion and noise, and
$f_{\mathrm{out}}(\cdot)$ captures output-side readout effects. The effective
MVM input after DAC conversion is modeled as
\begin{equation}
\label{eq:operand_noise_model}
\begin{aligned}
f_{\mathrm{DAC}}(\mathbf x)
=
\|\mathbf x\|_\infty
\left(
\frac{\mathbf x}{\|\mathbf x\|_\infty}
\odot(\mathbf 1+\mathbf m_{\mathbf x})
+\mathbf q_{\mathbf x}
\right) =
\mathbf x\odot(\mathbf 1+\mathbf m_{\mathbf x})
+\|\mathbf x\|_\infty\mathbf q_{\mathbf x} := \mathbf x+\boldsymbol\varepsilon_{{\mathbf x}}.
\end{aligned}
\end{equation}
Here $\odot$ denotes element-wise multiplication, $\mathbf 1$ is the
all-ones vector, and $\boldsymbol\varepsilon_{\mathbf x}$ is the total
input perturbation. 
The first expression makes the DAC-domain perturbation explicit: the input is first
normalized by $\|\mathbf x\|_\infty$ before entering the DAC, element-wise multiplicative noise
$\mathbf m_{\mathbf x}$ and quantization error $\mathbf q_{\mathbf x}$ are
applied in the normalized domain, and the result is then scaled back to the
original input range. For each input dimension $j \in \{1,\ldots,n\}$, the multiplicative noise is modeled as
\(
[\mathbf m_{\mathbf x}]_j
\overset{\mathrm{i.i.d.}}{\sim}
\mathcal N(0,\sigma_x^2),
\)
where $\sigma_x$ controls the input-noise level. The DAC quantization error is
modeled as
\(
[\mathbf q_{\mathbf x}]_j
\overset{\mathrm{i.i.d.}}{\sim}
\mathrm{Unif}
\left[
-\frac{1}{2^{b_{\mathrm{DAC}}}-2},
\frac{1}{2^{b_{\mathrm{DAC}}}-2}
\right],
\)
where $b_{\mathrm{DAC}}$ is the DAC precision in bits.
The same input-noise model is used for analog backward MVMs, with the
backpropagated error replacing $\mathbf x$. The programming non-idealities associated with updating the trainable LoRA
weights are described separately in Sec.~\ref{sec:update_stabilization}.

\subsection{Input reshaping}

We use a one-time pre-fine-tuning calibration to identify persistent
high-magnitude input channels, and use these statistics to reshape the inputs
seen by the low-precision analog paths. Input reshaping is applied differently
to the frozen base weights and the trainable LoRA weights.

For the frozen base path, we use function-preserving base-path rescaling.
Before fine-tuning, we estimate input-channel magnitudes from calibration
batches separately for each layer to which reshaping is applied. Let
$|[\widetilde{\mathbf x}]_i|$ denote the resulting magnitude statistic for
input channel $i=1,\ldots,n$. We construct a diagonal rescaling matrix
\(
  \mathbf D \propto
  \operatorname{diag}\!\left(
  |[\widetilde{\mathbf x}]_1|^\lambda,\ldots,
  |[\widetilde{\mathbf x}]_n|^\lambda
  \right),
\)
where $\lambda\in(0,1]$ controls the rescaling strength. The calibration
statistics and $\mathbf D$ remain fixed throughout fine-tuning. The input
and the corresponding frozen weights are rescaled together as
\begin{equation}
\label{eq:base_path_diag_rescaling}
    \mathbf W^{\mathrm{base}}\mathbf x
    =
    (\mathbf W^{\mathrm{base}}\mathbf D)
    (\mathbf D^{-1}\mathbf x),
    \qquad
    \overline{\mathbf W}^{\mathrm{base}}
    :=\mathbf W^{\mathrm{base}}\mathbf D,
    \qquad
    \overline{\mathbf x}:=\mathbf D^{-1}\mathbf x .
\end{equation}
Thus, the ideal base-path computation is unchanged, while persistent
high-magnitude input channels are reduced in
$\overline{\mathbf x}$. Because the base weights are frozen,
$\overline{\mathbf W}^{\mathrm{base}}$ is programmed to the analog array once
before fine-tuning.
The benefit follows directly from the DAC-noise model in
Eq.~\eqref{eq:operand_noise_model}. After rescaling, the analog input
perturbation is
\(
\boldsymbol\varepsilon_{\overline{\mathbf x}}
=
\overline{\mathbf x}\odot\mathbf m_{\overline{\mathbf x}}
+
\|\overline{\mathbf x}\|_\infty
\mathbf q_{\overline{\mathbf x}}\).
The rescaling is designed to reduce the input quantization-noise scale
$\|\overline{\mathbf x}\|_\infty$. The output-side effect also depends on
the rescaled weights; Figure~\ref{fig:noise_compression_diagnostics}
reports the resulting noise ratios.

For the trainable LoRA path, we use outlier-channel splitting. Using the same
calibration statistics, we select a persistent outlier-channel set
$\mathcal L$. The channels in $\mathcal L$ are computed on the protected path, while the remaining channels $\mathcal L^c=\{1,\ldots,n\}\setminus\mathcal L$ are kept on the
low-precision analog path. 
The vectors $\mathbf x_{\mathcal L}$ and $\mathbf x_{\mathcal L^c}$
contain the corresponding input channels, and
$\mathbf A_{\mathcal L}$ and $\mathbf A_{\mathcal L^c}$ contain the
corresponding columns of $\mathbf A$.
The $A$-side LoRA output is
\begin{equation}
\label{eq:adapter_path_split}
\mathbf z
=
\underbrace{
\mathbf A_{\mathcal L}\mathbf x_{\mathcal L}
}_{\text{protected analog path}}
+
\underbrace{
\mathbf A_{\mathcal L^c}
\left(
\mathbf x_{\mathcal L^c}\odot
(\mathbf 1+\mathbf m_{\mathcal L^c})
+
\|\mathbf x_{\mathcal L^c}\|_\infty
\mathbf q_{\mathcal L^c}
\right)
}_{\text{low-precision analog path}} .
\end{equation}
Removing the persistent outlier channels from the analog path replaces the DAC
range $\|\mathbf x\|_\infty$ by
$\|\mathbf x_{\mathcal L^c}\|_\infty$. The analog-input perturbation is therefore
\(
\boldsymbol\varepsilon_{\mathcal L^c}
=
\mathbf x_{\mathcal L^c}\odot\mathbf m_{\mathcal L^c}
+
\|\mathbf x_{\mathcal L^c}\|_\infty
\mathbf q_{\mathcal L^c}.
\)
Because the calibrated persistent outlier channels are excluded from the
analog path,
\(\|\mathbf x_{\mathcal L^c}\|_\infty \leq \|\mathbf x\|_\infty\),
with a strict reduction whenever a dominant input channel belongs to
\(\mathcal L\). This reduces the DAC-quantization term on the analog LoRA
path in those cases. The selected channels are computed on the protected
perfect-I/O analog path and are therefore not subject to the modeled input
perturbation in Eq.~\eqref{eq:operand_noise_model}.

\subsection{Update-path stabilization for analog LoRA}
\label{sec:update_stabilization}

The trainable LoRA weights are also affected by non-ideal physical
programming. Finite conductance resolution limits how small an update can be
programmed to the analog LoRA weights. We use digital mini-batch aggregation
and update accumulation \cite{nandakumar2020mixed}. Gradients are aggregated
digitally before the optimizer computes the weight update, and the same update
accumulation procedure is then applied to the resulting update regardless of
the optimizer choice. Updates smaller than the    conductance step are
retained until they can be programmed. 

Let $t$ denote an optimizer step and
$\boldsymbol\Theta_t\in\{\mathbf A_t,\mathbf B_t\}$ denote either trainable
LoRA weight matrix. Using a    uniform update resolution over $[-\tau,\tau]$, where $\tau$
is the weight-range bound, we set
$\Delta w_{\min}=2\tau/N$.
Let $\Delta\boldsymbol\Theta_t$ be the signed weight update produced by
the optimizer, instantiated as either Muon or AdamW in our experiments,
after digital mini-batch aggregation \cite{wu2026makinganalogtrainingscale}. A digital accumulator $\mathbf R_t$  then stores the optimizer update before it is
converted into an integer number of programming pulses:
\begin{align}
\mathbf R_{t+\frac12}
=
\mathbf R_t+\Delta\boldsymbol\Theta_t,\quad
\mathbf P_t
=
\operatorname{trunc}\!\left(
\mathbf R_{t+\frac12}/\Delta w_{\min}
\right),\quad
\mathbf R_{t+1}
=
\mathbf R_{t+\frac12}
-\Delta w_{\min}\mathbf P_t .
\end{align}
Here $\operatorname{trunc}(\cdot)$ is applied element-wise toward zero.  $\mathbf R_{t+\frac12}$ denotes the accumulator after adding the current
update, and $\mathbf P_t$ is the signed integer pulse-count matrix.

\section{Experiments}
\label{sec:experiments}

We evaluate robust analog LoRA fine-tuning using IBM Analog Hardware
Acceleration Kit (AIHWKit)~\cite{buchel2024aihwkit}. We consider three main
testbeds spanning question answering (QA) and instruction tuning: the Llama-3-8B, Llama-3.2-1B-Instruct, and
Qwen2.5-7B-Instruct.
For the two QA testbeds, we use an equal mixture of NQ-Open, TriviaQA, and
WebQuestions
\cite{kwiatkowski2019natural,lee2019latent,joshi2017triviaqa,
berant2013semantic}. For Llama-3-8B, the Q/K/V/O projections and
corresponding LoRA weights in the first six decoder blocks are mapped to
analog arrays; for Llama-3.2-1B-Instruct, the corresponding projections in
all 16 decoder blocks are mapped to analog arrays. Training uses 2,000
optimizer steps, sequence length 256, LoRA rank $r=8$, and scaling factor
$\alpha=16$. Input reshaping is applied to Q/K/V using one-time
pre-fine-tuning calibration over 64 batches, with $\lambda=0.8$ for
base-path rescaling and the eight most persistent high-magnitude channels
selected for outlier-channel splitting. The primary analog configuration
uses    $N=100$, 7-bit input DACs, and input-noise standard deviation
0.02 in both forward and backward MVMs. Perfect-I/O controls remove these
modeled MVM input perturbations. We report F1 and Exact Match (EM) using
greedy generation on a fixed 512-example QA set, together with
response-token negative log-likelihood (NLL) on 100 held-out examples.

For instruction tuning, we fine-tune
\texttt{Qwen2.5-7B-Instruct} on \texttt{OpenThoughts-114k} using the Qwen
chat template~\cite{guha2026openthoughts}. The Q/K/V projections and
corresponding LoRA weights in the last six decoder blocks are mapped to
analog arrays. Input reshaping uses fixed pre-fine-tuning calibration with
$\lambda=0.8$ and eight persistent high-magnitude input channels, with the
selected channels routed through the same protected perfect-I/O analog path
used in the primary experiments. We evaluate held-out assistant-response-token
NLL across multiple conductance-resolution and DAC/noise settings. Additional implementation details and
supplementary testbeds are provided in Appendix~\ref{app:settings}.

\subsection{Question-Answering Results}
\label{sec:qa_results}

The QA experiments address two complementary questions. The first two result
columns in Table~\ref{tab:qa_main} evaluate input reshaping. Under noisy MVM
computation, reshaping improves F1 and EM for both models and both optimizers,
while its effect is small under perfect I/O, where the targeted input-side
perturbations are removed. The Llama-3.2-1B-Instruct results further show that
the benefit persists when analog mapping is extended to all 16 decoder blocks.

The last two result columns in Table~\ref{tab:qa_main}   isolate update accumulation while keeping
reshaping enabled. Discarding the programming residual substantially degrades
F1 and EM for both models and optimizers, demonstrating the importance of
retaining sub-threshold updates under finite-resolution programming.
Validation NLL in Table~\ref{tab:qa_nll} shows the same qualitative trends
for both input reshaping and update accumulation. The    $N=20$ stress
test in Table~\ref{tab:softmax_qa_n20} further shows that both benefits
persist under coarser programming resolution.
\begin{table}[t]
\centering
\caption{\textbf{Primary QA results.}
Each entry reports F1/EM (\%). Reshaping denotes base-path rescaling + outlier-channel splitting.
UA denotes update accumulation.
I/O $b/s$ denotes a
$b$-bit input DAC with input-noise standard deviation $s$. }
\label{tab:qa_main}

\small
\setlength{\tabcolsep}{4pt}
\renewcommand{\arraystretch}{1.06}

\begin{tabular}{@{}lllccc@{}}
\toprule
Model & I/O & Optimizer &
\makecell{No reshaping\\+ UA} &
\makecell{Reshaping\\+ UA} &
\makecell{Reshaping\\w/o UA} \\
\midrule

Llama-3-8B & Perfect & Muon
& 46.84 / 37.89 & 47.84 / 39.26 & 35.51 / 26.17 \\

Llama-3-8B & Perfect & AdamW
& 48.50 / 40.63 & 48.36 / 40.43 & 35.51 / 26.17 \\

\midrule

Llama-3-8B & 7/0.02 & Muon
& 24.49 / 18.36 & \textbf{43.55 / 35.35} & 28.73 / 19.92 \\

Llama-3-8B & 7/0.02 & AdamW
& 31.51 / 24.22 & \textbf{43.86 / 35.94} & 30.32 / 20.90 \\

\midrule

Llama-3.2-1B & Perfect & Muon
& 30.88 / 23.24 & 30.10 / 24.22 & 10.61 / 0.00 \\

Llama-3.2-1B & Perfect & AdamW
& 28.19 / 21.09 & 29.34 / 21.88 & 10.61 / 0.00  \\

\midrule

Llama-3.2-1B & 7/0.02 & Muon
& 19.21 / 12.89 & \textbf{25.24 / 17.97} & 8.92 / 0.00 \\

Llama-3.2-1B & 7/0.02 & AdamW
& 16.48 / 11.13 & \textbf{23.70 / 16.80} & 9.02 / 0.00 \\

\bottomrule
\end{tabular}
\end{table}

\subsection{Instruction-Tuning Results}
\label{sec:instruction_results}
Table~\ref{tab:qwen} evaluates input reshaping on Qwen2.5-7B-Instruct
across multiple noise settings and two    conductance resolutions.
Reshaping lowers mean held-out NLL in all eight noisy within-optimizer comparisons for both Muon and AdamW.
Together with the Llama QA results, these results show that input reshaping remains effective across different models, tasks, optimizers, and analog settings.

\begin{table}[!htbp]
\centering
\caption{\textbf{Qwen/OpenThoughts held-out NLL.}
Mean $\pm$ standard deviation over three seeds; lower is better.
UA is enabled throughout.}
\label{tab:qwen}
\small
\setlength{\tabcolsep}{3.2pt}
\begin{tabular}{@{}c l c c c c@{}}
\toprule
  $N$ & I/O setting
& \multicolumn{2}{c}{Muon [NLL$\downarrow$]}
& \multicolumn{2}{c}{AdamW [NLL$\downarrow$]} \\
\cmidrule(lr){3-4}\cmidrule(lr){5-6}
& & { No reshaping  } & Reshaping  
& { No reshaping  } & Reshaping   \\
\midrule
-- & Digital
& 0.6823{\scriptsize$\pm$0.0003} & --
& 0.7235{\scriptsize$\pm$0.0001} & -- \\

100 & Perfect
& 0.7121{\scriptsize$\pm$0.0006} & --
& 0.7548{\scriptsize$\pm$0.0013} & -- \\

20 & Perfect
& 0.7896{\scriptsize$\pm$0.0025} & --
& 0.8727{\scriptsize$\pm$0.0000} & -- \\

100 & 4/0.02
& 0.7356{\scriptsize$\pm$0.0009}
& \textbf{0.7209}{\scriptsize$\pm$0.0001}
& 0.7852{\scriptsize$\pm$0.0005}
& \textbf{0.7649}{\scriptsize$\pm$0.0007} \\

100 & 6/0.01
& 0.7154{\scriptsize$\pm$0.0004}
& \textbf{0.7104}{\scriptsize$\pm$0.0002}
& 0.7569{\scriptsize$\pm$0.0007}
& \textbf{0.7521}{\scriptsize$\pm$0.0005} \\

20 & 4/0.02
& 0.8158{\scriptsize$\pm$0.0019}
& \textbf{0.7952}{\scriptsize$\pm$0.0010}
& 0.9128{\scriptsize$\pm$0.0001}
& \textbf{0.8908}{\scriptsize$\pm$0.0004} \\

20 & 5/0.03
& 0.8051{\scriptsize$\pm$0.0029}
& \textbf{0.7911}{\scriptsize$\pm$0.0018}
& 0.8936{\scriptsize$\pm$0.0005}
& \textbf{0.8812}{\scriptsize$\pm$0.0002} \\
\bottomrule
\end{tabular}
\end{table}
\subsection{Training-Time versus Evaluation-Only Reshaping}
\label{sec:eval_only_results}

\begin{table}[t]
\centering
\caption{\textbf{Training-time versus evaluation-only reshaping.}
F1 (\%) is reported as mean $\pm$ sample standard deviation over
three repeated noisy evaluations. Without adaptation, reshaping only uses
 base-path rescaling (BR). With adaptation, reshaping uses both BR and  outlier-channel splitting.}
\label{tab:eval_only_br_os}

\small
\setlength{\tabcolsep}{4.5pt}
\renewcommand{\arraystretch}{1.08}

\begin{tabular}{@{}llccc@{}}
\toprule
Setting & Optimizer & No reshaping & Eval-only reshaping& Train+eval reshaping\\
\midrule
No adaptation
& --    & $7.12 \pm 0.07$  & $9.14 \pm 0.09$  & -- \\

\multirow{2}{*}{Adaptation }
& AdamW & $18.20 \pm 1.06$ & $22.02 \pm 0.26$ & $\mathbf{23.57 \pm 0.28}$ \\
& Muon  & $19.52 \pm 0.73$ & $21.82 \pm 0.67$ & $\mathbf{24.41 \pm 0.58}$ \\
\bottomrule
\end{tabular}
\end{table}

Motivated by analog-inference methods such as
NORA~\cite{hou2025nora}, Table~\ref{tab:eval_only_br_os} examines
whether input reshaping is useful only as an inference-time correction,
or whether applying it during fine-tuning provides an additional benefit.
We use Llama-3.2-1B-Instruct under the primary noisy-I/O setting
($N=100$, 7-bit input DAC, and input noise $0.02$), with three repeated
noisy evaluations on the same 512 QA examples.
Without adaptation, applying function-preserving base-path
rescaling only at inference already improves the F1 score of frozen model. 
We next evaluate the same question after LoRA adaptation, where input
reshaping combines function-preserving base-path rescaling and
outlier-channel splitting. With analog LoRA adaptation, reshaping already improves F1 when applied
only at evaluation, from $18.20$ to $22.02$ for AdamW and from $19.52$
to $21.82$ for Muon. Applying reshaping during both fine-tuning and
evaluation further increases F1 to $23.57$ and $24.41$, respectively.
Together, these results show that input reshaping benefits both inference and analog LoRA fine-tuning, while the large improvement over the frozen-model result ($9.14$ F1 score) demonstrates effective task adaptation under noisy analog computation.

The supplementary experiments further examine generalization and the individual
components of the method. On Llama-3-8B/OpenOrca,
Table~\ref{tab:openorca} reports lower mean response-token NLL with
reshaping across all noisy settings tested. The recovered linear-attention
Llama-3-8B experiments in Tables~\ref{tab:qa_100_full}
and~\ref{tab:qa_20_full} provide an additional architecture variant, with
F1 and EM improving in most noisy comparisons across multiple conductance
resolutions.
We also provide diagnostics to better understand the reshaping mechanism.
Table~\ref{tab:reshaping_ablation} evaluates the individual contributions
of base-path rescaling and outlier-channel splitting.
Figure~\ref{fig:qwen_calibration_stability} examines whether the channels
selected during one-time calibration remain persistent during fine-tuning.
Finally, Fig.~\ref{fig:noise_compression_diagnostics} reports analytical
noise ratios associated with the two reshaping components. Together, these supplementary results complement the main-task results with additional task and architecture coverage, component ablations, and mechanism-level diagnostics.

\section{Conclusion and Limitations}
We present a parameter-efficient approach for adapting pretrained LLMs to
analog hardware without full-model hardware-aware retraining. Input reshaping
mitigates input-side MVM errors, while update accumulation preserves
sub-threshold updates under finite-resolution programming. Across the primary
QA and additional evaluations, these techniques improve analog LoRA
fine-tuning under the simulated non-idealities considered. Our study is limited to AIHWKit simulation and selected models and device
settings; it does not measure physical-device behavior or end-to-end latency,
energy, programming cost, or communication overhead. Some supplementary
stress-test and component-ablation results are point estimates. Physical
validation, broader device coverage, and system-level evaluation remain
future work.

\clearpage
\bibliographystyle{IEEEtranN}
\bibliography{odi_references}

\clearpage
\appendix
\begin{center}
  {\LARGE\bfseries Supplementary Material}
\end{center}

\section{Experimental Details}
\label{app:settings}

We use IBM AIHWKit~\cite{rasch2021flexible,le2023using} to simulate analog
LoRA fine-tuning. Pretrained weights remain frozen, while the LoRA weights
in the mapped layers are trainable. DAC precision and input noise are
applied to both forward and backward analog MVMs. Digital mini-batch
aggregation is performed before the Muon or AdamW optimizer step, followed
by the update-accumulation procedure described in
Sec.~\ref{sec:update_stabilization}.
Throughout the paper, \emph{No reshaping + UA} disables input reshaping
while retaining update accumulation (UA), \emph{Reshaping + UA} enables
both components, and \emph{Reshaping w/o UA} enables reshaping but discards
the programming residual after each transfer. 
Perfect-I/O analog controls remove the modeled MVM input perturbations
while retaining the analog weight representation and programming procedure. I/O $b/s$ denotes a
$b$-bit input DAC with input-noise standard deviation $s$.

\subsection{Primary Llama QA testbeds}
\label{app:primary_llama_settings}

Both primary testbeds use an equal mixture of NQ-Open, TriviaQA, and
WebQuestions. Fine-tuning uses 2,000 optimizer steps, sequence length 256,
micro-batch size one, digital aggregation over eight micro-batches, LoRA
rank $r=8$, and scaling factor $\alpha=16$. The base learning-rate amplitude
is $10^{-4}$ and decays to zero. Muon uses momentum 0.95, while AdamW uses
$(\beta_1,\beta_2)=(0.9,0.999)$ and weight decay 0.01.
The corresponding Q/K/V/O projections and LoRA weights are mapped to analog
arrays, while input reshaping is applied to Q/K/V. Calibration uses 64
pre-fine-tuning batches, $\lambda=0.8$, and the eight channels most
frequently attaining the largest absolute token activation. The calibration
statistics and selected channels remain fixed during fine-tuning.
The primary noisy-I/O setting uses 7-bit input DACs and input-noise standard
deviation 0.02.

Evaluation uses greedy generation on a fixed 512-question QA set with
standard answer normalization and best matching over gold-answer aliases.
Validation NLL is computed on 100 held-out examples using response-token
negative log-likelihood.

\paragraph{Llama-3-8B.}
For the original softmax-attention Llama-3-8B, the Q/K/V/O projections and
corresponding LoRA weights in the first six decoder blocks are mapped to
analog arrays. The primary experiments use   $N=100$.
For the $N=20$ stress test, AdamW uses a fivefold learning-rate multiplier
relative to the $N=100$ setting. 

\paragraph{Llama-3.2-1B-Instruct.}
For Llama-3.2-1B-Instruct, Q/K/V/O projections and their corresponding LoRA
weights in all 16 decoder blocks are mapped to analog arrays. MLPs,
embeddings, normalization layers, attention softmax, and the language-model
head remain digital. The analog device uses   $N=100$ with programming
increment $\Delta w_{\min}=0.02$.

\subsection{Additional testbeds}
\paragraph{Llama/OpenOrca.}
We fine-tune the softmax-attention Llama-3-8B on OpenOrca using an
instruction--response format. The last eight decoder blocks' Q/K/V
projections and corresponding LoRA weights are mapped to analog arrays.
Input reshaping uses $\lambda=0.8$ and eight calibrated outlier channels.
Evaluation reports final response-token NLL. The evaluated analog settings
are given in Table~\ref{tab:openorca}.

\paragraph{Recovered linear-attention Llama QA.}
We additionally evaluate a linear-attention Llama-3-8B checkpoint recovered through
LoLCATs~\cite{grattafiori2024llama,zhang2024lolcats} on the same three QA
datasets. The first six decoder blocks' Q/K/V/O projections and corresponding
LoRA weights are mapped to analog arrays, with reshaping applied to Q/K/V.

\section{Additional Results}

\subsection{Additional QA Results}
\label{app:primary_qa_results}

Table~\ref{tab:qa_nll} reports validation NLL for the primary QA
experiments and shows the same qualitative trends as F1/EM:
reshaping improves noisy-I/O performance, while discarding the programming
residual substantially degrades performance. Table~\ref{tab:softmax_qa_n20}
extends the evaluation to    $N=20$, where both reshaping and update
accumulation remain beneficial.

\begin{table}[!htbp]
\centering
\caption{\textbf{Validation NLL for the primary QA experiments.}
Lower is better.}
\label{tab:qa_nll}

\small
\setlength{\tabcolsep}{5pt}
\renewcommand{\arraystretch}{1.06}

\begin{tabular}{@{}lllccc@{}}
\toprule
Model & I/O & Optimizer &
\multicolumn{3}{c}{Validation NLL} \\
\cmidrule(lr){4-6}
& & &
\makecell{No reshaping + UA} &
\makecell{Reshaping + UA} &
\makecell{Reshaping w/o UA} \\
\midrule

Llama-3-8B
& Perfect
& Muon
& 1.8349
& 1.8250
& -- \\

Llama-3-8B
& Perfect
& AdamW
& 1.8728
& 1.8650
& -- \\

\midrule

Llama-3-8B
& 7/0.02
& Muon
& 2.3346
& \textbf{1.9617}
& 4.5610 \\

Llama-3-8B
& 7/0.02
& AdamW
& 2.5149
& \textbf{2.1374}
& 4.4824 \\

\midrule

Llama-3.2-1B
& 7/0.02
& Muon
& 3.0193
& \textbf{2.9235}
& 6.7572 \\

Llama-3.2-1B
& 7/0.02
& AdamW
& 3.1864
& \textbf{3.0138}
& 6.3773 \\

\bottomrule
\end{tabular}
\end{table}

\begin{table}[!htbp]
\centering
\caption{\textbf{Llama-3-8B QA at $N=20$.}
F1 and EM are in \%. Results are point estimates.}
\label{tab:softmax_qa_n20}

\small
\setlength{\tabcolsep}{4pt}
\renewcommand{\arraystretch}{1.05}

\begin{tabular}{@{}l
ccc
ccc@{}}
\toprule
&
\multicolumn{3}{c}{Muon} &
\multicolumn{3}{c}{AdamW} \\
\cmidrule(lr){2-4}
\cmidrule(lr){5-7}

Configuration &
F1 &
EM &
NLL &
F1 &
EM &
NLL \\
\midrule

No reshaping + UA
& 18.45 & 12.70 & 3.1689
& 26.89 & 20.31 & 3.0256 \\

Reshaping + UA
& \textbf{39.96} & \textbf{31.84} & \textbf{2.3762}
& \textbf{39.41} & \textbf{31.84} & \textbf{2.1780} \\

Reshaping w/o UA
& 31.08 & 21.29 & 4.4745
& 30.26 & 20.70 & 4.5044 \\

\bottomrule
\end{tabular}
\end{table}

\subsection{Additional Instruction-Tuning Results}
\label{app:additional_results}

Table~\ref{tab:openorca} extends the instruction-tuning evaluation to
Llama-3-8B/OpenOrca. Reshaping lowers mean response-token NLL in all six
noisy within-optimizer comparisons, consistent with the Qwen/OpenThoughts
results in Table~\ref{tab:qwen}.

\begin{table}[!htbp]
\centering
\caption{\textbf{Llama-3-8B/OpenOrca response-token NLL.}
Mean $\pm$ standard deviation over three seeds; lower is better.
UA is enabled throughout.}
\label{tab:openorca}
\small
\setlength{\tabcolsep}{3.2pt}
\begin{tabular}{@{}c l c c c c@{}}
\toprule
  $N$ & I/O setting
& \multicolumn{2}{c}{Muon [NLL$\downarrow$]}
& \multicolumn{2}{c}{AdamW [NLL$\downarrow$]} \\
\cmidrule(lr){3-4}\cmidrule(lr){5-6}
& & \makecell{No reshaping  } & Reshaping  
& \makecell{No reshaping  } & Reshaping   \\
\midrule
$10^4$ & Perfect
& 0.6841{\scriptsize$\pm$0.0024} & --
& 0.7001{\scriptsize$\pm$0.0008} & -- \\

100 & Perfect
& 0.7164{\scriptsize$\pm$0.0012} & --
& 0.7373{\scriptsize$\pm$0.0014} & -- \\

100 & 4/0.01
& 0.7355{\scriptsize$\pm$0.0021}
& \textbf{0.7239}{\scriptsize$\pm$0.0015}
& 0.7564{\scriptsize$\pm$0.0025}
& \textbf{0.7429}{\scriptsize$\pm$0.0024} \\

100 & 6/0.02
& 0.7302{\scriptsize$\pm$0.0021}
& \textbf{0.7211}{\scriptsize$\pm$0.0019}
& 0.7488{\scriptsize$\pm$0.0024}
& \textbf{0.7419}{\scriptsize$\pm$0.0023} \\

$10^4$ & 6/0.02
& 0.7048{\scriptsize$\pm$0.0015}
& \textbf{0.6941}{\scriptsize$\pm$0.0001}
& 0.7151{\scriptsize$\pm$0.0023}
& \textbf{0.7070}{\scriptsize$\pm$0.0039} \\
\bottomrule
\end{tabular}
\end{table}

\subsection{Recovered Linear-Attention QA Results}
\label{app:qa_full}

Tables~\ref{tab:qa_100_full} and~\ref{tab:qa_20_full} report the complete
recovered-QA results. F1 and EM each improve in seven of eight noisy
comparisons. The exceptions are AdamW EM at $N=100$, 6/0.01, and AdamW
F1 at $N=20$, 6/0.01; both metrics improve together in six comparisons.

\begin{table}[!htbp]
\centering
\caption{\textbf{Recovered linear-attention Llama-3-8B QA.}
Results at $N\in\{100,10^4\}$ are mean $\pm$ standard deviation
over three seeds.}
\label{tab:qa_100_full}
\small
\setlength{\tabcolsep}{3.5pt}
\begin{tabular}{@{}r l c c c c@{}}
\toprule
  $N$ & I/O
& \multicolumn{2}{c}{Muon}
& \multicolumn{2}{c}{AdamW} \\
\cmidrule(lr){3-4}\cmidrule(lr){5-6}
& & No reshaping + UA & Reshaping + UA & No reshaping + UA & Reshaping + UA \\
\midrule
$10^4$ & Perfect
& \qametric{51.58}{1.4}{44.33}{1.4}
& --
& \qametric{51.84}{0.7}{44.53}{1.1}
& -- \\

100 & Perfect
& \qametric{50.67}{1.2}{43.69}{0.5}
& --
& \qametric{49.95}{0.9}{42.64}{0.6}
& -- \\

100 & 7/0.02
& \qametric{46.84}{1.3}{39.45}{1.3}
& \qametric{50.81}{0.3}{43.49}{0.8}
& \qametric{45.00}{0.4}{38.35}{0.9}
& \qametric{49.06}{0.2}{41.73}{0.5} \\

100 & 6/0.01
& \qametric{49.76}{1.6}{42.06}{1.4}
& \qametric{50.72}{0.2}{43.29}{0.2}
& \qametric{48.42}{0.5}{41.60}{0.6}
& \qametric{48.85}{0.8}{41.15}{0.2} \\
\bottomrule
\end{tabular}
\end{table}

\begin{table}[!htbp]
\centering
\caption{\textbf{Recovered linear-attention Llama-3-8B QA at $N=20$.}
Results are point estimates.}
\label{tab:qa_20_full}
\small
\setlength{\tabcolsep}{4pt}
\begin{tabular}{@{}r l c c c c@{}}
\toprule
  $N$ & I/O
& \multicolumn{2}{c}{Muon}
& \multicolumn{2}{c}{AdamW} \\
\cmidrule(lr){3-4}
\cmidrule(lr){5-6}
& & No reshaping + UA & Reshaping + UA & No reshaping + UA & Reshaping + UA \\
\midrule
20 & Perfect
& 49.01 / 42.38
& --
& 50.00 / 43.16
& -- \\

20 & 7/0.02
& 40.42 / 33.40
& 48.08 / 41.41
& 45.85 / 38.09
& 49.64 / 42.19 \\

20 & 6/0.01
& 47.54 / 41.21
& 48.77 / 41.80
& 48.20 / 40.62
& 47.95 / 41.02 \\
\bottomrule
\end{tabular}
\end{table}

\section{Input-Reshaping Ablations and Diagnostics}
\label{app:calibration}

\subsection{Reshaping ablations}

Table~\ref{tab:reshaping_ablation} compares base-path rescaling alone
with full reshaping. Base-path rescaling provides most of the observed improvement; adding outlier-channel splitting further improves both F1 and EM.

\begin{table}[!htbp]
\centering
\caption{\textbf{Input-reshaping component ablation.}
Llama-3-8B QA with Muon under    $N=100$ and I/O setting 7/0.02.
Results are point estimates of F1 (EM), in \%. BR denotes function-preserving base-path rescaling and OS denotes
outlier-channel splitting.}
\label{tab:reshaping_ablation}
\small
\setlength{\tabcolsep}{8pt}
\begin{tabular}{@{}ccc@{}}
\toprule
BR Off, OS Off &
BR On, OS Off &
BR On, OS On \\
\midrule
24.49 (18.36) &
42.01 (33.79) &
\textbf{43.55 (35.35)} \\
\bottomrule
\end{tabular}
\end{table}

\subsection{Calibration stability}

\begin{figure}[!htbp]
\centering
\includegraphics[width=0.5\linewidth]{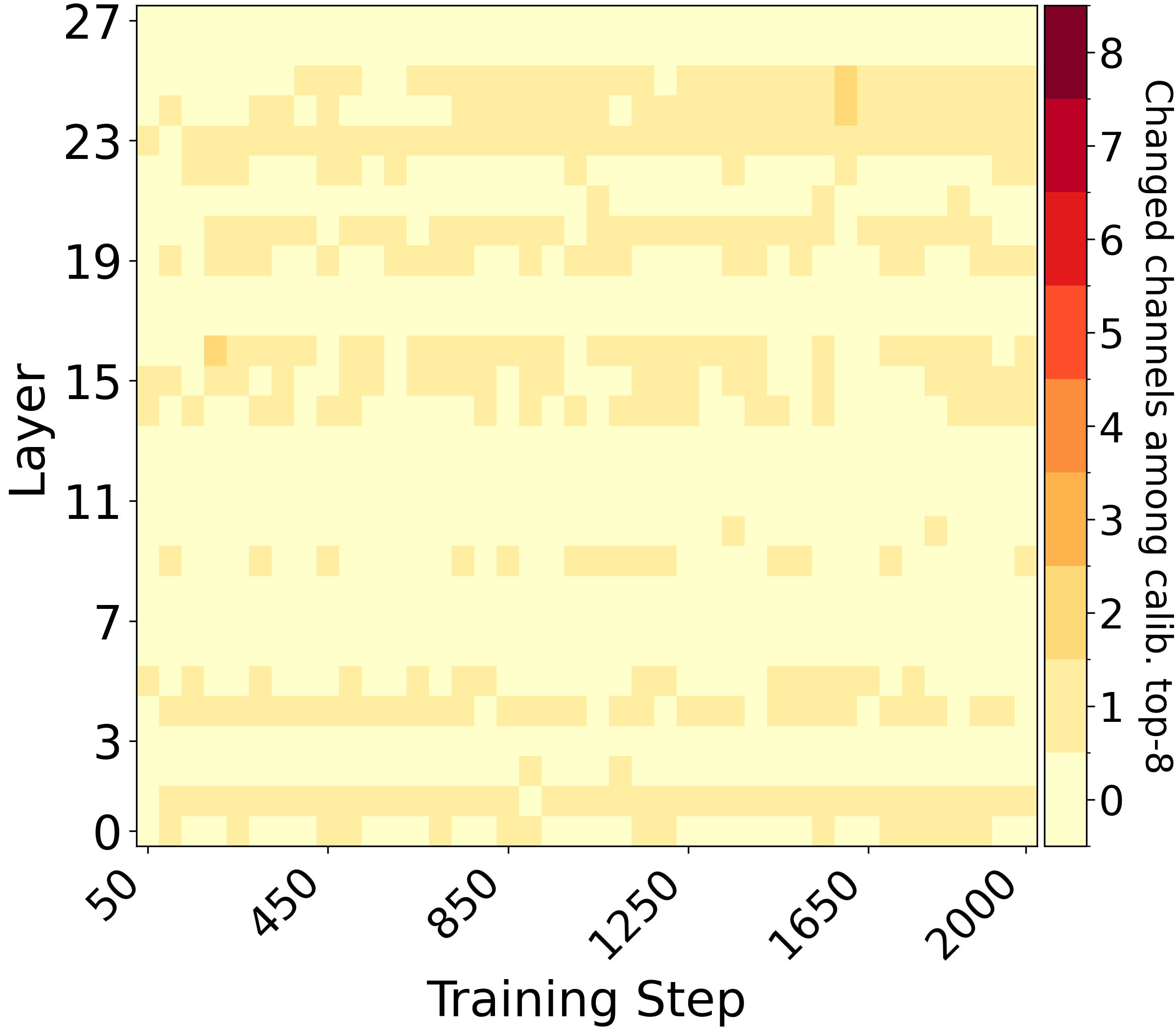}
\caption{\textbf{Calibration stability on Qwen/OpenThoughts.}
Each cell counts channels in a 64-step window's top-8 set that are absent
from the calibration set; zero indicates full overlap. Rows are monitored
layers and columns are training windows.}
\label{fig:qwen_calibration_stability}
\end{figure}

Figure~\ref{fig:qwen_calibration_stability} compares the fixed
pre-fine-tuning top-8 outlier set with rankings recomputed over 64-step
training windows. Channels are ranked by how often they attain the
largest absolute token activation. Most selected channels remain shared
with the calibration set over the plotted interval.

\subsection{Quantization-noise compression}

\begin{figure}[!htbp]
\centering
\begin{minipage}[t]{0.42\linewidth}
  \centering
  \includegraphics[width=\linewidth]{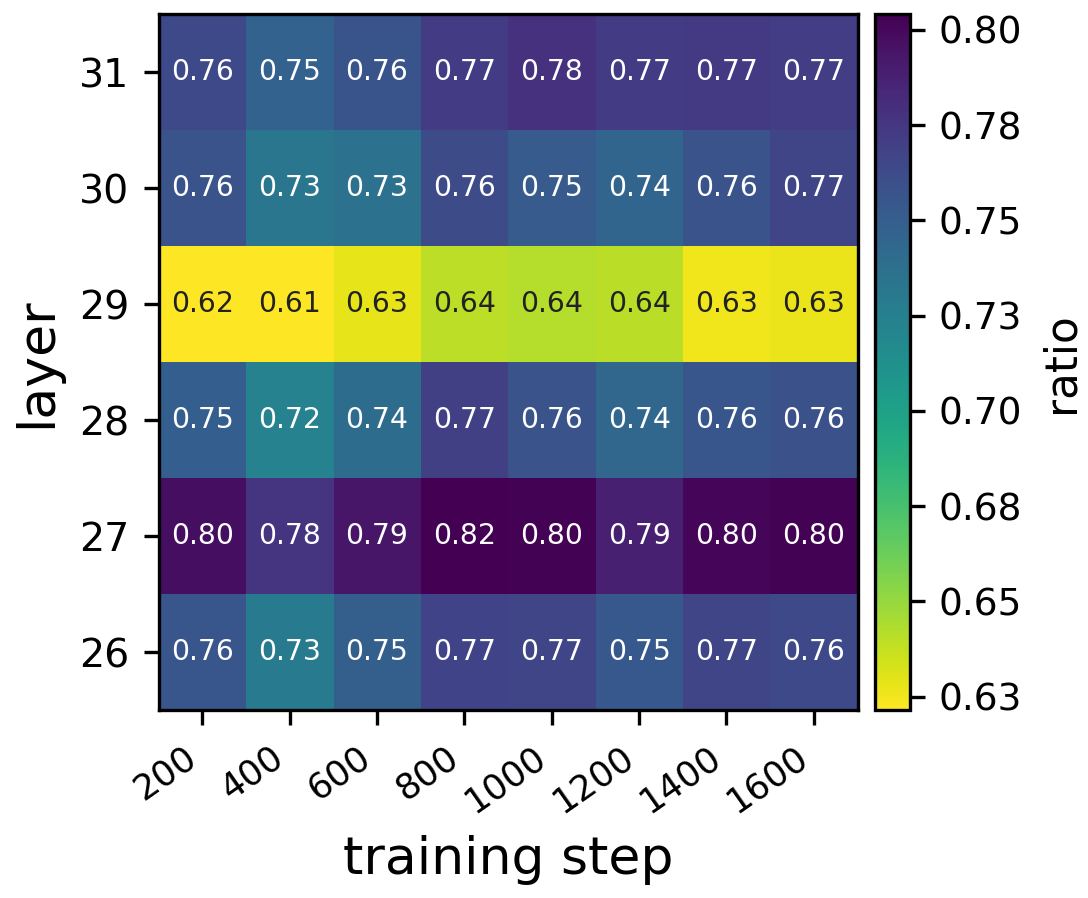}
\end{minipage}
\hfill
\begin{minipage}[t]{0.42\linewidth}
  \centering
  \includegraphics[width=\linewidth]{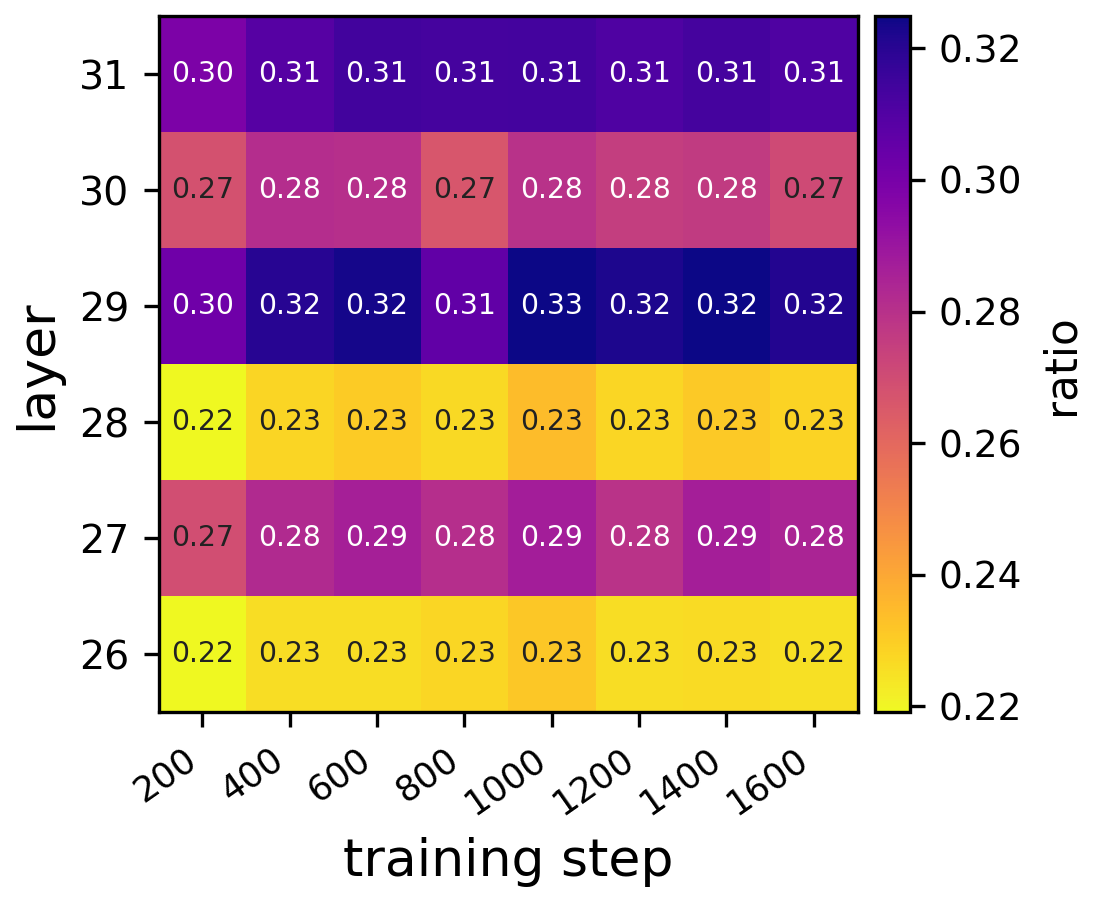}
\end{minipage}
\caption{\textbf{Input-reshaping quantization-noise diagnostics.}
Left: remaining analog channels' input-noise variance ratio
(Eq.~\eqref{eq:supp_split_noise_ratio}). Right: base-path output-noise
energy ratio, including weight rescaling
(Eq.~\eqref{eq:supp_base_noise_ratio}). Cells average per-token ratios
within 200-step windows; values below one indicate compression.}
\label{fig:noise_compression_diagnostics}
\end{figure}

Figure~\ref{fig:noise_compression_diagnostics} reports analytical ratios
for the additive DAC quantization-noise component in
Eq.~\eqref{eq:operand_noise_model}, computed from recorded inputs and
weights. Each cell averages per-token ratios within a 200-step window.
For outlier-channel splitting, the calibrated channels $\mathcal L$
use the protected path. At fixed DAC precision, the per-channel
quantization-noise variance ratio on the remaining analog inputs is
\begin{equation}
\label{eq:supp_split_noise_ratio}
\left(
\frac{\|\mathbf x_{\mathcal L^c}\|_\infty}
     {\|\mathbf x\|_\infty}
\right)^2.
\end{equation}
For base-path rescaling, let
$\overline{\mathbf x}=\mathbf D^{-1}\mathbf x$ and
$\overline{\mathbf W}^{\mathrm{base}}=\mathbf W^{\mathrm{base}}\mathbf D$.
Under independent, equal-variance additive quantization errors, the
modeled output-noise energy ratio is
\begin{equation}
\label{eq:supp_base_noise_ratio}
C_y(\mathbf x)
=
\left(
\frac{\|\overline{\mathbf x}\|_\infty}
     {\|\mathbf x\|_\infty}
\right)^2
\frac{\|\overline{\mathbf W}^{\mathrm{base}}\|_F^2}
     {\|\mathbf W^{\mathrm{base}}\|_F^2},
\end{equation}
where $\|\cdot\|_F$ is the Frobenius norm. The two factors account for
changes in input range and weight norm. The plotted ratios are
approximately 0.61--0.82 for splitting and 0.22--0.33 for base-path
rescaling, remaining below one throughout the displayed windows.

\end{document}